\documentclass[letterpaper, 10pt, conference]{ieeeconf}
\IEEEoverridecommandlockouts                              

\usepackage{graphics} 
\usepackage{graphicx} 
\graphicspath{ {Figures/}}
\usepackage{epsfig} 
\usepackage{mathptmx} 
\usepackage{times} 
\usepackage{amsmath} 
\usepackage{amssymb}  
\usepackage{wasysym}
\usepackage{siunitx}
\usepackage{morefloats}
\usepackage[noadjust]{cite}

\usepackage{color,soul}
\usepackage{textcomp}
\usepackage{array}
\usepackage{makecell}
\usepackage{tabularx}
\usepackage{tabulary}
\usepackage{subfigure}

\usepackage{lipsum}

\usepackage{amsmath}
\usepackage{amssymb}
\usepackage{latexsym}
\usepackage{url}
\usepackage{cite}
\usepackage{relsize}
\usepackage{multirow}
\usepackage{afterpage}
\usepackage{ifthen}
\usepackage{graphicx}
\usepackage{algpseudocode,algorithm,algorithmicx}
\usepackage{subfigure}
\usepackage[keeplastbox]{flushend}
\usepackage{epstopdf}
\usepackage{stackengine}
\usepackage{gensymb}
\usepackage[bookmarks=false, linkcolor=blue, urlcolor=blue, citecolor=blue]{hyperref} 
\usepackage{booktabs}
\usepackage{makecell}
\usepackage{hhline}
\usepackage{courier}
\usepackage{lipsum}
\usepackage{xcolor}
\usepackage{bm}

\usepackage{siunitx}

\newcommand{\ra}[1]{\renewcommand{\arraystretch}{#1}}

\hypersetup{pdfstartview={FitH}}

\title{\LARGE \bf Nonlinearity Compensation in A Multi-DoF Shoulder Sensing Exosuit For Real-Time Teleoperation
}

\author{
Rejin John Varghese$^{1}$, Anh Nguyen$^{1}$, Etienne Burdet$^{2}$, Guang-Zhong Yang$^{3}$ and Benny P L Lo$^{1}$
\thanks{This research is supported by the Future AI and Robotics for Space (FAIR-SPACE) grant (EP/R026092/1) awarded by the EPSRC, UK.} 
\thanks{$^{1}$The Hamlyn Centre, Department of Computing, Institute for Global Health Innovation, Imperial College London, UK (email: 
	{\tt\small r.varghese15@imperial.ac.uk})}%
\thanks{$^{2}$Department of Bioengineering, Imperial College London, UK}
\thanks{$^{3}$Institute of Medical Robotics, Shanghai Jiao Tong University, China}
}

\begin{document}
\maketitle
\thispagestyle{empty}
\pagestyle{empty}

\begin{abstract}

The compliant nature of soft wearable robots makes them ideal for complex multiple degrees of freedom (DoF) joints, but also introduce additional structural nonlinearities. Intuitive control of these wearable robots requires  robust sensing to overcome the inherent nonlinearities. This paper presents a joint kinematics estimator for a bio-inspired multi-DoF shoulder exosuit capable of compensating the encountered nonlinearities. To overcome the nonlinearities and hysteresis inherent to the soft and compliant nature of the suit, we developed a deep learning-based method to map the sensor data to the joint space. 
The experimental results show that the new learning-based framework outperforms recent state-of-the-art methods by a large margin while achieving $12ms$ inference time using only a GPU-based edge-computing device. The effectiveness of our combined exosuit and learning framework is demonstrated through real-time teleoperation with a simulated NAO humanoid robot.

\end{abstract}
\section{INTRODUCTION}
\label{sec:intro}

As opposed to rigid-bodied systems, soft exoskeletons (exosuits) leverage the body's anatomical structures to form the robot's frame and use unorthodox materials and actuation methods, thereby yielding improved compliance, reduced weight and profile, lower power requirements, and improved affordability \cite{Varghese2018,Asbeck2014a}. In the recent past, soft exoskeleton robots from research groups and industry have  demonstrated their capability to successfully provide partial lower- and upper-limb rehabilitation and assistance \cite{Varghese2018,Maciejasz2014,Asbeck2014a,Polygerinos2015a,Galiana2012,Lessard2017}. The compliance inherently present in exosuits make them ideal for assisting complex multiple degrees of freedom (DoF) joints such as the shoulder that do not behave kinematically as conventional ball-and-socket joints \cite{Varghese2018,Li2018,Perry2007,Nef2009,lessard2018soft,Gaponov2017}.

\begin{figure}[t]
\centering
\includegraphics[width=\linewidth]{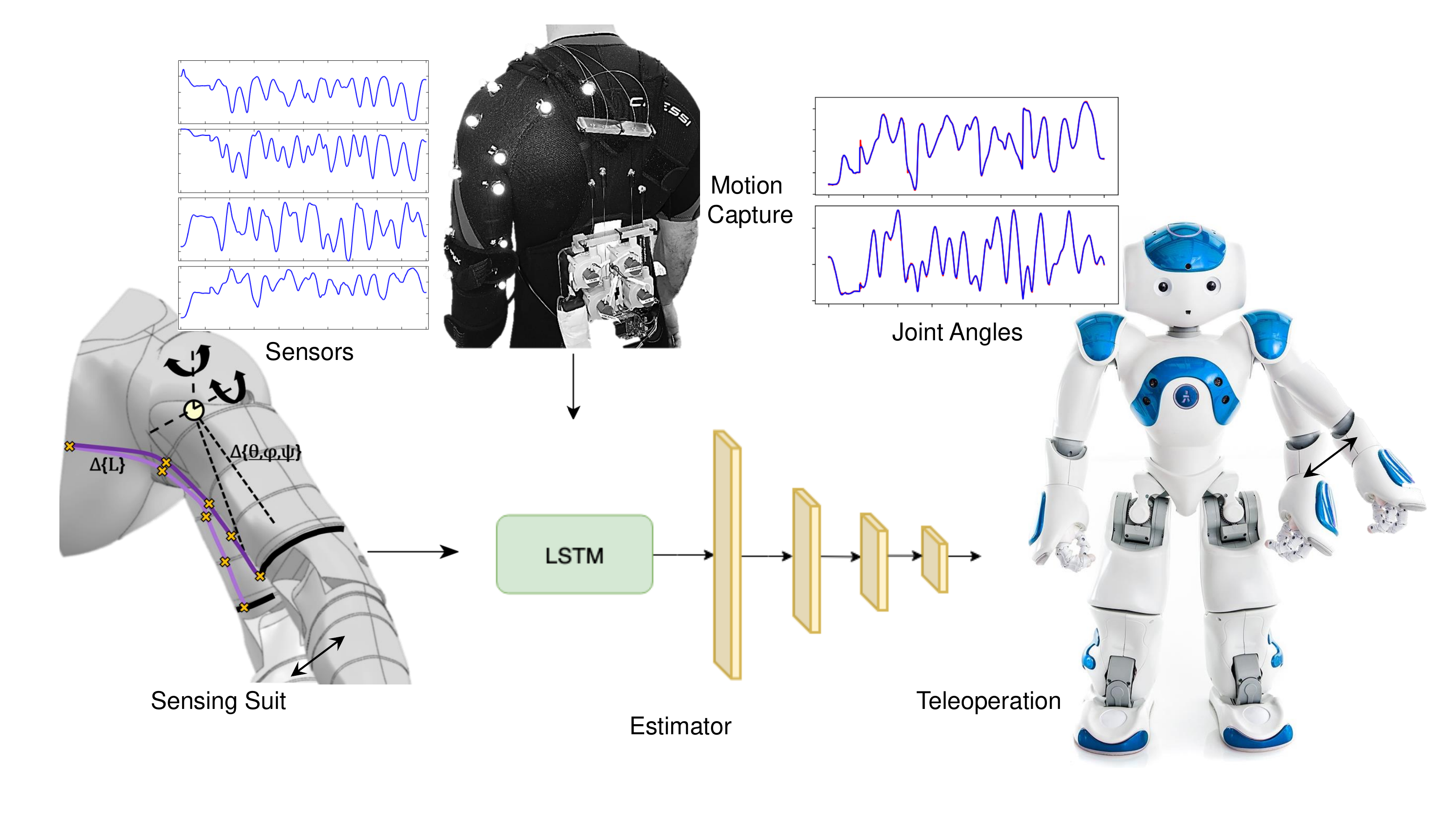}

\caption{Overview of the proposed sensor fusion and nonlinearity compensation framework in a multi-DoF sensing suit for real-time teleoperation. A schematic of the sensing framework concept is shown wherein a change in the upper-limb pose is tracked by 4 sensors, from which the joint kinematics are estimated. The estimated angles may then be used to teleoperate a robot.}
\label{fig:overallSchematic}
\end{figure}

For exosuits to be capable of providing transparent, intuitive and natural assist-as-needed control, robust kinematic sensing frameworks that behave analogously to the body's sense of proprioception are essential. 
Traditional encoder-based joint angle sensing is not possible in soft-bodied suits, especially at the body's multi-DoF joints. Researchers have explored alternative traditional sensing modalities such as Inertial Measurement Units (IMUs) \cite{Aslani2018,Asbeck2014a}, bend/flex sensors \cite{Galiana2012}, and string-pot sensors \cite{varghese2019design}, and also newer smart-material-based sensing modalities such as printed circuit fabrics \cite{nakajima2019wiring}, liquid metal alloys \cite{Menguc2014}, and micro-fluidics \cite{Chossat2015} to determine the body's kinematics. Off the newer soft-sensing systems, only \cite{varghese2019design} and \cite{nakajima2019wiring} are wearable garment-like systems, and allow full range of motion in multiple DoFs. While soft-bodied technologies provide excellent compliance and mechanical transparency, this advantage is also the technology's Achilles heel, as this compliant soft nature introduces additional layers of nonlinearities. Soft-sensing systems encounter nonlinearities such as gradual unrecoverable deformation (creep) \cite{nakajima2019wiring,varghese2019design,Chossat2015,Menguc2014} and hysteresis \cite{nakajima2019wiring,Menguc2014,Chossat2015,varghese2019design}, leading to sub-optimal sensing capabilities when compared to the rigid-bodied equivalents. 

Recently, two wearable suits for sensing the upper-limb joint kinematics were proposed in \cite{varghese2019design} and \cite{nakajima2019wiring}. To map the sensor outputs from the suits to the joint space, both \cite{varghese2019design} and \cite{nakajima2019wiring,ogata2019estimating} used learning-based techniques with neural networks. The soft nature of these suits introduce nonlinearities like hysteresis and creep, and hence, the learning methods employed to perform the multivariate multiple regression from sensor-to-joint space need to play an additional role of being able to compensate the encountered nonlinearities. However, the learning-based mapping solutions described in \cite{varghese2019design} and \cite{ogata2019estimating} are unable to compensate the nonlinearities accurately, and result in a higher root mean square error (RMSE) compared to the state-of-the-art sensor-only approach (i.e., IMU \cite{Aslani2018}) in estimating upper-limb joint kinematics.

In this work, we present a learning-based framework capable of sensor fusion and nonlinearity compensation to overcome the aforementioned shortcoming in the sensing framework presented in \cite{varghese2019design}. Two equivalent sensing systems, one using string-pot sensors and another by generating splines from motion-capture (mocap) data were presented in \cite{varghese2019design}. We show that by combining our new learning technique with the spline-based sensing framework, we can effectively compensate the inherent nonlinearities and outperform the IMU-based solution in estimating shoulder kinematics in 2 DoFs. Finally, we demonstrate the robustness of the entire framework through real-time teleoperation with a humanoid robot in simulation. A graphical overview of the proposed framework is presented in Fig.\ref{fig:overallSchematic}. 


Next, we present the state-of-the-art in kinematic sensing technologies and nonlinearity estimation techniques for different systems. In Section \ref{sec:sensFrmWk}, we briefly present the sensing framework of our exosuit, and the experimental setup and methodology for nonlinearity quantification. In Section \ref{sec:sensorMapping}, we introduce our learning method and the methodology to perform the real-time teleoperation, and this is followed by a presentation and discussion of the results in Section \ref{sec:expRes}. Finally, we conclude the paper and discuss future work in Section \ref{sec:conc}.

\section{RELATED WORK}
\label{sec:relWork}


Motion-capture and RGB-camera-based pose estimation technologies have limited applications due to their dependence on sensors (\textit{e.g.} cameras) placed extrinsic to the wearer. Self-contained wearable kinematic systems could have applications such as continuous health monitoring, control and teleoperation of robots and exoskeletons, and applications in augmented/virtual reality (AR/VR) and gaming. Commercial and research IMU-based products are considered the state-of-the-art in wearable kinematics sensing systems, but suffer from drift which can become significant without periodic correction \cite{Varghese2018,ogata2019estimating}. Recently, researchers have investigated both traditional \cite{Galiana2012,varghese2019design} and newer smart-material-based \cite{Menguc2014,nakajima2019wiring} sensing technologies for wearable kinematic sensing applications. Due to the soft and compliant nature of both the embodiment on which these sensors are mounted, and the smart-material-based sensing systems, nonlinearities such as hysteresis and gradual permanent deformation (creep) have been observed, and need to be corrected to ensure accurate joint kinematics tracking \cite{varghese2019design,ogata2019estimating}. 

Material deformation-induced nonlinearity estimation and compensation has been studied extensively in piezoelectric \cite{ru2009hysteresis} and shape memory alloy-based \cite{liu2010tracking} actuators and force/torque sensors \cite{koike2019hysteresis}, but very few solutions have been developed for wearable kinematic sensing systems. Nonlinearity estimation has been researched for smart-material-based sensing systems such as tactile (artificial skin) \cite{urban2015sensor}, strain \cite{liu2010tracking}, and both textile- \cite{meyer2010design} and elastomer- \cite{chuan2011application} based pressure sensors. Researchers have investigated both model-based \cite{ru2009hysteresis,koike2019hysteresis} and data-driven \cite{urban2015sensor,chuan2011application} nonlinearity estimation techniques. Some examples of hysteresis modelling methods include the Preisach \cite{visone2008hysteresis}, Prandtl-Ishilinksii \cite{ru2009hysteresis}, and Duhem \cite{liu2010tracking} models. Different data and learning-based methods such as Gaussian processes \cite{urban2015sensor}, neural network \cite{islam2006hysteresis} and Support Vector Machine (SVM) \cite{chuan2011application,wang2008hysteresis} have also been employed for hysteresis estimation. 

Though total nonlinearity experienced by an actuator or sensor is rate-dependent, most research on nonlinearity estimation either ignore or approximate the rate dependent nonlinearity as a first order function \cite{koike2019hysteresis}. This assumption, however, does not hold true when significant constant loading is present as is the case in the sensing suit presented in \cite{varghese2019design}. In \cite{wang2008hysteresis}, the rate-dependent total nonlinearity was estimated by periodically retraining the SVM. In \cite{koike2019hysteresis}, time-series information (current and data from one previous step) were fed as two inputs to an neural network to estimate hysteresis, but temporal data was not leveraged to estimate rate-dependent nonlinearity. The e-skin wearable sensing suit, a commercially available product developed by Xenoma Inc. and presented in \cite{nakajima2019wiring,ogata2019estimating}, uses fabric-based strain-sensing to estimate joint kinematics. The authors in \cite{ogata2019estimating} estimate joint kinematics using the e-skin suit \cite{nakajima2019wiring} by inputting a time-series of sensor data into a Convolutional Neural Network (CNN). Though, the e-skin suit is a commercially available suit with a high Technology Readiness Level (TRL), the neural network proposed in \cite{ogata2019estimating} was unable to compensate the nonlinearities with very high accuracy and exhibited poorer performance than IMU-based sensing technologies \cite{Aslani2018} in tracking multi-DoF upper-limb joint kinematics.

The sources of nonlinearities in the sensing framework introduced by the authors previously in \cite{varghese2019design} and detailed in Section \ref{subsec:nonlinearities}, differ significantly from other smart-material-based kinematic sensing systems. In the sensing frameworks presented in \cite{Menguc2014,Chossat2015}, the nonlinear behaviour is inherent in the sensing modality itself. Conversely, in the suit proposed in \cite{varghese2019design}, it is not the sensor but the fabric embodiment and skin together (during the coupled 2 DoF movements) that introduce the hysteretic behaviour. Additionally, the constant spring force exerted by the string-pot sensor on the garment and other skin-suit interactions induce creep-like rate-dependent nonlinearities as well. 
Long Short-Term Memory (LSTM) networks \cite{hochreiter1997long} can encode both spatial and temporal information and, therefore, be used to learn the temporal dependence of the above-mentioned nonlinearities. The research presented in \cite{varghese2019simulation} showed that LSTM performed superior to vanilla neural network in forecasting multi-DoF joint kinematics from a data stream of multiple sensors. We, therefore, explore the use of LSTM-based sensor-to-joint-space mappings in this work to perform sensor fusion and encode the rate-dependent total nonlinearity to achieve more accurate joint kinematics tracking.

\section{SENSING SUIT DESIGN AND NONLINEARITY QUANTIFICATION}
\label{sec:sensFrmWk}

\subsection{Design Concept and Prototype}
\label{subsec:SuitConcept}

\begin{figure}
\centering
\includegraphics[width = 0.9\linewidth]{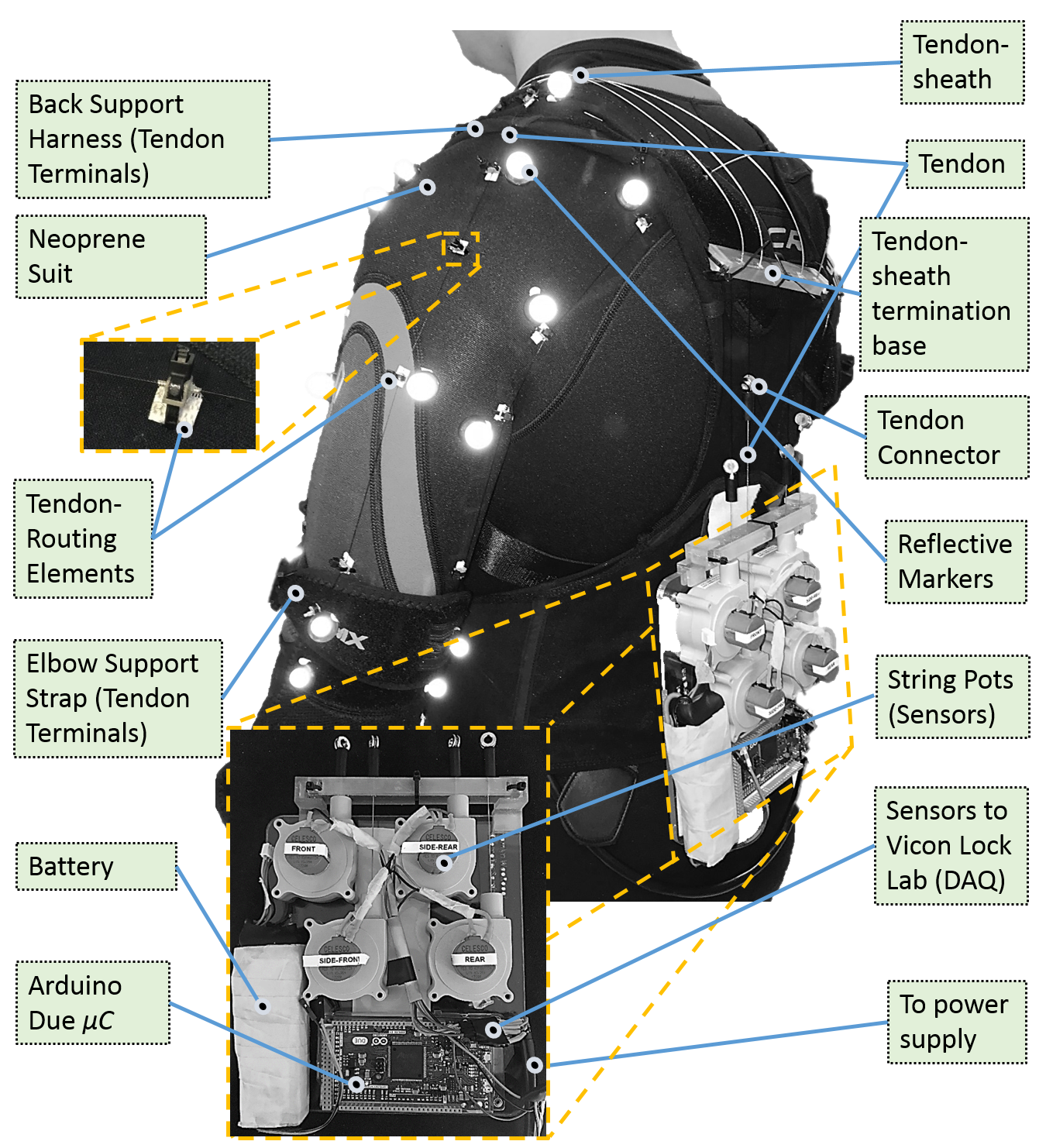}
\caption{Overview of the sensing framework prototype for the shoulder exosuit with a description of the electro-mechanical and mechanical elements (figure adapted from \cite{varghese2019design}). The nonlinearity compensation in this work uses the analogous virtual spline-fitted sensing framework}
\label{fig:hardwareModel}
\end{figure}

As an exosuit uses the human-body as the frame of the robot, traditional encoder-based joint kinematics sensing is not possible. Our exosuit's sensing framework \cite{varghese2019design}, is inspired by the body's sense of proprioception,  to get sense of a limb's position and movement \cite{Proske2009,Winter2005}. 
To reduce the number of tendons required for 2 DoF ((azimuth ($\theta$) and elevation ($\phi$)) sensing, a dimensionality reduction method motivated by the muscle synergy concept is adopted \cite{Bernstein1967}.  
This concept states that movement control by the Central Nervous System is done through multiple muscles that work together as a synergy, rather than individually. At the shoulder, it has been shown that 5+ muscles work together in 4 synergies to generate movements in the 2 DoFs \cite{Budhota2017,d2013control}. A similar muscle synergy-inspired dimensionality reduction is employed in \cite{varghese2019design}, and resulted in a framework made of 4 tendon-based sensing units- $\mathbb{F,\text{ }SF, \text{ }SR,\text{ }R}$ placed on the front, side-front, side-rear and rear side of the shoulder respectively \cite{varghese2019design}. The tendons are routed along specific paths on the suit and changes in their path lengths are recorded as the arm moves (Fig.\ref{fig:overallSchematic} and \ref{fig:hardwareModel}). Sensor fusion from multiple sensors is then performed to derive the joint kinematics in 2 DoFs.

To test the above-discussed hypothesis, the suit shown in Fig.\ref{fig:hardwareModel} was developed, which is presented in detail in \cite{varghese2019design}. The suit consists of a form-fitting neoprene embodiment on which four tendons are routed using multiple 3D-printed routing elements. 
One end of the tendon is fixed above the elbow on the upper-arm, and the free end is connected to a string potentiometer to measure its displacement. The spring in the sensor applies a small constant force ($\sim1.9\si{\newton}$)
which allows the tendon length to be tracked accurately without slacking
The potentiometer in the sensor is connected to an Arduino Due (Arduino, Italy) microcontroller ($\mu$C) over a voltage divider circuit, which allows for displacement to be measured as an analog signal. 
The Arduino $\mu$C is then connected to a Jetson Nano (NVidia, CA, USA) embedded system-on-module board. The program running on the Arduino streams sensor data to the Jetson Nano at a rate of $250\si{\hertz}$. The Jetson Nano consists of a 128-core NVidia Maxwell Graphics Processing Unit (GPU) and allows for the learning-based sensor-to-joint-space mapping to decode the shoulder joint kinematics from the incoming sensor data stream.
The electronics were mounted on the back support making the entire system portable (Fig.\ref{fig:hardwareModel}). The total weight of the electromechanical components and the base is approximately $1.1\si{\kilo\gram}$. The prototype presented in this work only consists of a sensing framework and is the test-platform to finalise the tendon-routing architecture before attempting actuation. For more details on the bio-inspired design concept and developed hardware, please refer to \cite{varghese2019design}. 

We would like to point out that a virtual spline-based sensing framework was employed for nonlinearity compensation in this work (detailed in following sub-sections). This was done as the current prototype of the sensing suit had physical stops that were introduced to expose the tendons for improved visual demonstrability of the suit principle (see accompanying video). These physical stops limit the real sensors from moving outside a particular range of motion introducing other non-soft material related nonlinearities that will be addressed in a future version of the suit.

\subsection{Motion Capture Experiments for Nonlinearity Quantification}
\label{subsec:expSetup}
To characterise and benchmark the suit, motion capture (MoCap) experiments were performed. The experiments were performed with the ethics approval from the Imperial College Research Ethics Committee (ICREC reference number 18IC4816). A lab comprising of 10 Vero 2.2 cameras, a Vue video camera, a Lock Lab and ancillary equipment from Vicon (Vicon, Oxford, UK) were used for experiments. 
The subject wore a prototype with reflective markers attached to the anatomical features and 3-D printed routing elements. 
The string-pot sensors were connected directly to the Vicon Lock Lab to synchronise sensor and MoCap data acquisition. Sensor and camera data were obtained at $1200\si{\hertz}$ and $120\si{\hertz}$, respectively. A variety of movements in both azimuth and elevation DoFs were performed to evaluate the suit, and the details of the movements performed during MoCap experiments are presented in Table \ref{tab:exp_mvmts}. Over 29,500 frames of data was collected from these experiments.

\newcolumntype{K}[1]{>{\centering\arraybackslash}m{#1}}
\begin{table}[t]
\renewcommand{\arraystretch}{1.2}
\caption{Movements Performed by Subject During MoCap Experiments}
\label{tab:exp_mvmts}
\centering
 
 
\begin{tabular}{K{2.15cm}||K{2.2cm}|K{1cm}|K{1cm}}
\hline
\bfseries Movement & \bfseries Joint Angles Range/Limits & \bfseries No. of Reps. & \bfseries No. of Frames \\
\hline\hline
\bfseries Flexion-Extension & \makecell{$\theta={-90}^{0}/{90}^{0}$ \\ ${0}^{0}\leq\phi\leq{90}^0 $} & 4 & 3037 \\
 
\hline
\bfseries Abduction-Adduction & \makecell{$\theta={0}^{0}$ \\ ${0}^{0}\leq\phi\leq{90}^0$} & 4 & 3630  \\
 
\hline
\bfseries \makecell{Az. angle (fix.) \\ El. angle (var.)} & \makecell{$\theta=5$ const. vals. \\ ${0}^{0}\leq\phi\leq{90}^0$} & 2 & 5814  \\
\hline
\bfseries \makecell{El. angle (fix.) \\ Az. angle (var.)} & \makecell{ ${-40}^{0}\leq\theta\leq{90}^0$ \\ $\phi=5$ const. vals.} & 2 & 6757 \\
\hline
\bfseries Random movements & \makecell{${-40}^{0}\leq\theta\leq{90}^0$ \\ ${0}^{0}\leq\phi\leq{90}^0$} & N/A & 10313 \\
\hline
\multicolumn{3}{c}{\textbf{Total Frames of Mocap Expts.}} & 29551  \\
\hline
\end{tabular}
\end{table}

\begin{figure*}[t]
\centering
\includegraphics[width = 0.9\linewidth]{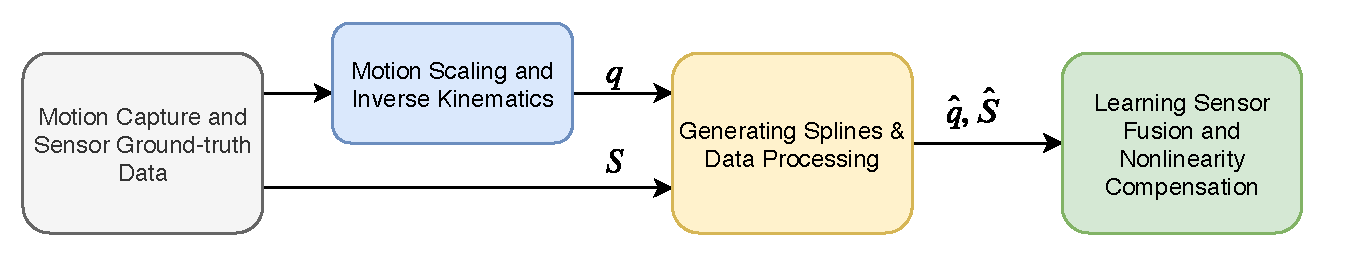}
\caption{Overview of the experimental methodology for nonlinearity quantification: Data from motion capture experiments are processed to derive the joint angle (from inverse kinematics in OpenSim) and real and spline-generated virtual sensor data. This data is then used to train the networks to compensate nonlinearities and predict the joint angles from 4 sensor values.}
\label{fig:expMethodSchematic}
\end{figure*}

\subsection{Motion Capture Data Processing}
\label{subsec:expMethod}
 
 

MoCap data from the markers affixed to the anatomical features on the skin (and not the suit) were used to derive ground-truth shoulder joint kinematics allowing us to decouple the ground-truth joint angle data source from the input data for the splines which was obtained solely from markers on the suit. Joint angle data was obtained by performing inverse kinematics computations on upper-limb musculoskeletal (MS) model in OpenSim \cite{Delp2007}, post-scaling the generic MS model to the subject's measurements. The generic upper-limb model used for this analysis was the MoBL-ARMS Dynamic Upper Limb model \cite{Saul2015}. The marker data was transformed from the Vicon reference frame to the OpenSim model reference frame in MATLAB (Mathworks, MA, USA) before performing the inverse kinematics computations.
As was presented in \cite{varghese2019design}, the 3D-printed routing elements were also tracked in the MoCap experiments as they govern the behaviour of the sensing tendons. Virtual sensors were derived using splines defined by joining these markers for each data frame (Fig.\ref{fig:hardwareModel}) \cite{varghese2019design}. The spline-generated virtual sensors are used as the sensing modality in this work for attempting nonlinearity compensation and teleoperation. An overview of the data processing methodology for quantifying sensing suit performance and nonlinearities is presented in Fig.\ref{fig:expMethodSchematic}.


\subsection{Observed Hysteresis and Creep Nonlinearities}
\label{subsec:nonlinearities}

\begin{figure}
\centering
\includegraphics[width = 0.9\linewidth]{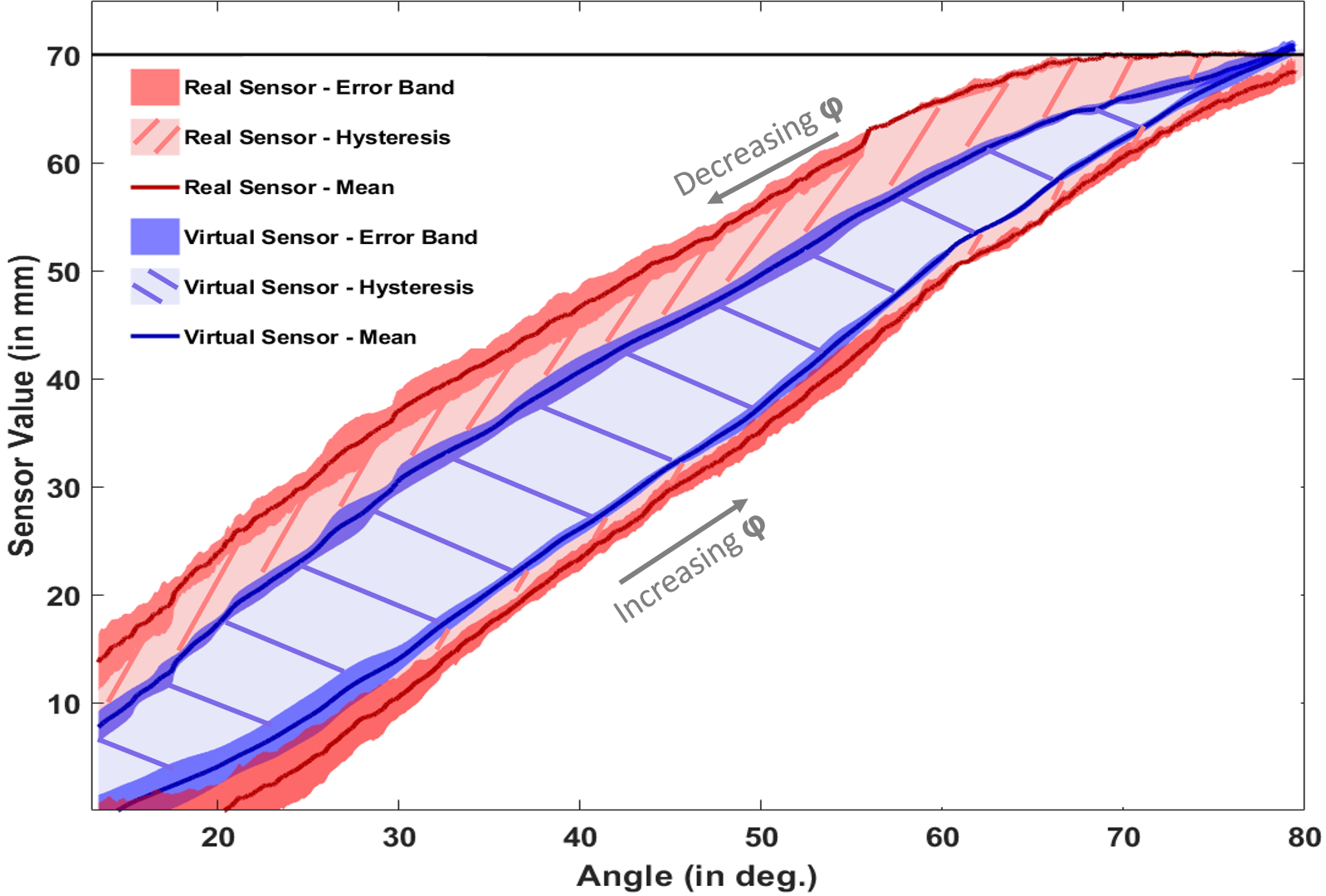}
\caption{Comparison of real and spline-generated sensor performance: The graph shows the evolution of real (red) and spline-generated (blue) sensor-$\mathbb{SF}$ values against elevation angle during multiple F/E movements. This graph presents the hysteresis and error bands observed in both the real and spline-generated sensors (figure adapted from \cite{varghese2019design}).}
\label{fig:noninearityRes}
\end{figure}

Fig.\ref{fig:noninearityRes} shows the change in a sensor value (sensor $\mathbb{SF}$) with changing elevation angle ($\phi$) during flexion-extension movements. Hysteresis behaviour is observed in both the real and spline-generated sensors.  The spring  forces  from  the  string-pots  acting  on  the  suit-fabric, the highly coupled 2 DoF movements at the shoulder, and the differences in skin and fabric deformation properties together result in both hysteresis- and creep-like nonlinearities. Hysteresis in the spline-generated sensors is observed primarily due to the inherent soft nature of the neoprene base suit on which the markers/routing-elements are mounted. The small permanent deformation is observed because of the constant spring force applied by the string-pots. This behaviour is further exacerbated, in the case of the real sensors due to limits imposed by the termination base (see Fig.\ref{fig:hardwareModel} and \ref{fig:noninearityRes}). The sensor is not able to recover completely from the imposed sensor limit due to: 1). friction between tendons and the routing elements, tendon-sheaths, \textit{etc.}, and 2). compliance experienced by the suit due to the spring force from the string-pot sensor. The permanent drift observed at the end of the movement cycle is due to the rate-dependent nonlinearity that builds up over time. Similar behaviour is observed in all the tendons during other movements as well. 

\begin{figure*}[!t]
  \centering
    \subfigure[LNNet]{\includegraphics[width=0.49\linewidth, height=0.25\linewidth]{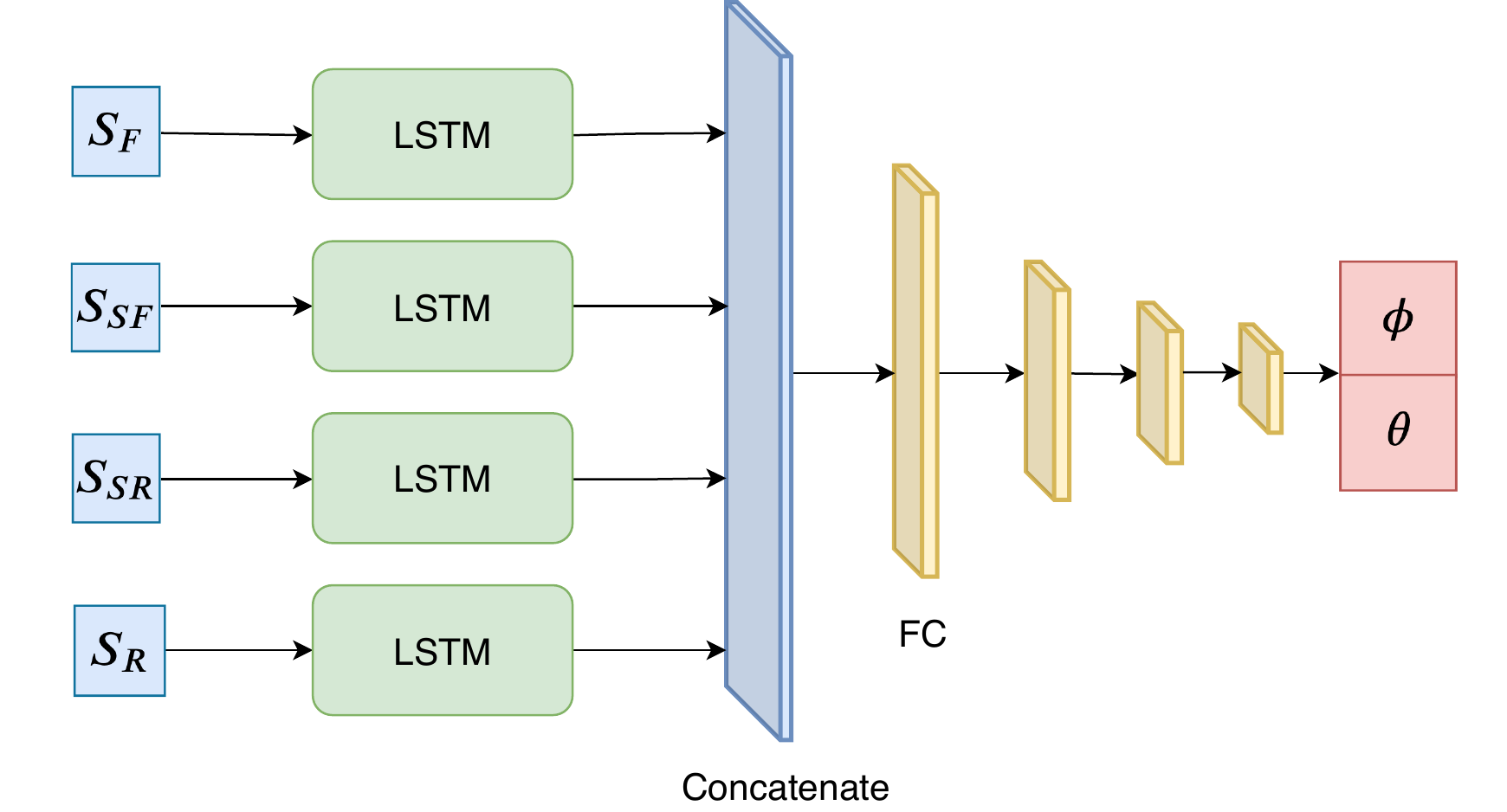}}
    \subfigure[ENNet]{\includegraphics[width=0.49\linewidth, height=0.25\linewidth]{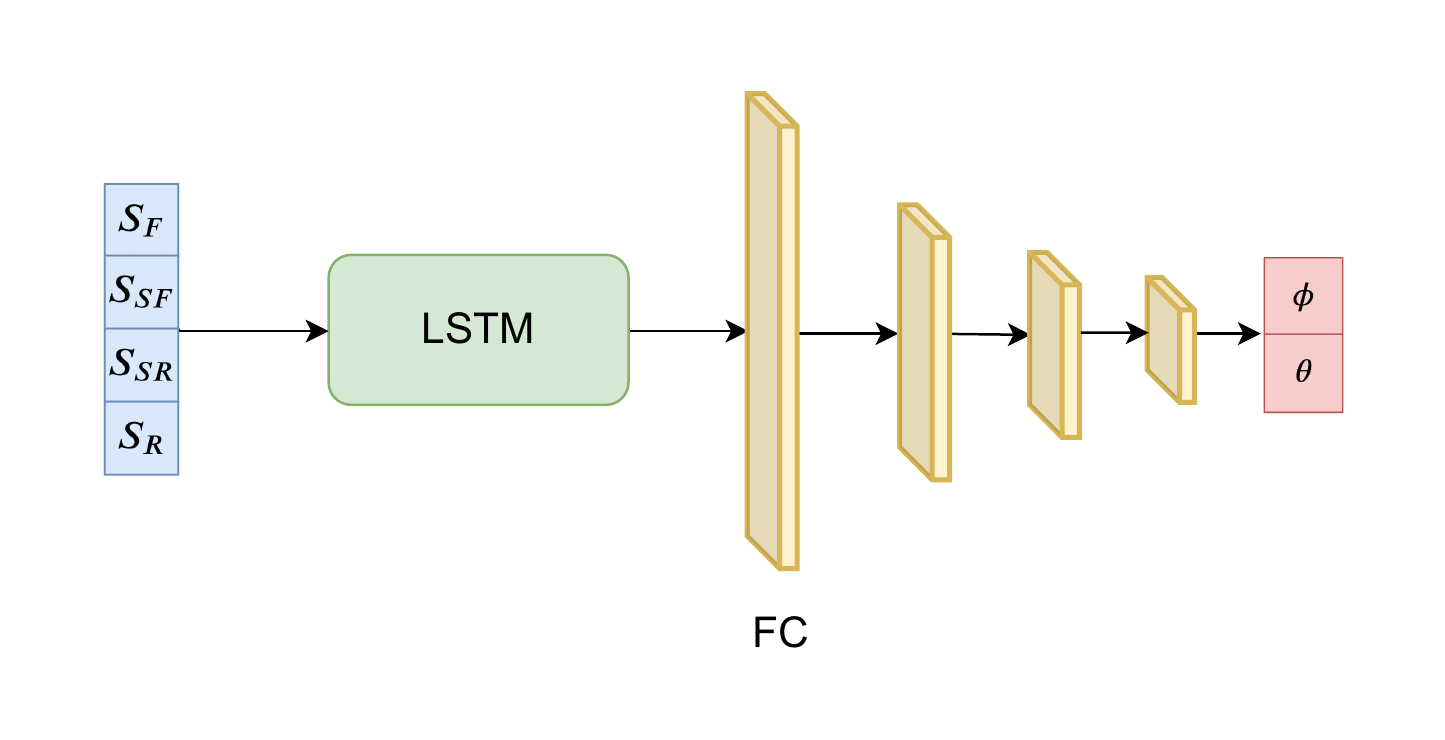}}
    \vspace{2ex}
 \caption{An overview of our networks: (a) The Late-fusion Nonlinearity Net (LNNet), and (b) The Early-fusion Nonlinearity Net (ENNet). The key difference between LNNet and ENNet is the fusion technique to handle the input sensor values before feeding to the LSTM network.}
 \label{Fig_learning_overview}
\end{figure*}

\section{SENSOR-TO-JOINT-SPACE MAPPING AND TELEOPERATION}
 \label{sec:sensorMapping}

\subsection{Learning}
\label{subsec:learning}

As discussed in the previous sections, in the case of our multi-DoF sensing framework, both rate-dependent (creep) and rate-independent (hysteresis) nonlinearities are intrinsically present, and need real-time online compensation. To this end, we propose new methods to learn the sensor-to-joint-space mapping between spline-generated virtual sensor data $\mathbf{S} = \{S_{F}, S_{SF}, S_{SR}, S_{R}\} \in \mathbb{R}^4$ and the output angles $\mathbf{q}=\{\theta, \phi\} \in \mathbb{R}^2 $. As the sensors generate the continuous data in sequential order, learning the long-term dependency from the data is the key to success. Mathematically, we can represent this temporal dependency and sensor fusion with the following equation:

\begin{equation}
q^{i} = f(S^{i-(n-1)},...,S^{i-1},S^{i}) \\
\end{equation}
where $S^{i} = [S_{F}^{i},S_{SF}^{i},S_{SR}^{i},S_{R}^{i}]$, $q^{i} = [\theta^{i},\phi^{i}]$ and a window of previous $n$ time steps are considered for forecasting joint kinematics at the $i^{th}$ time step. In this work, we propose two deep-learning-based methods for effectively encoding the temporal information from the raw sensory data. Both of our methods use LSTM neural networks \cite{hochreiter1997long} capable of learning long-term dependencies from the input data in many problems \cite{nguyen2019object,nguyen2019v2cnet}. A detailed illustration of our proposed methods can be found in Fig.\ref{Fig_learning_overview}.

In particular, in the first network design (i.e., Late-fusion Nonlinearity Net (LNNet) in Fig.~\ref{Fig_learning_overview}. a), we consider the input from four sensors are independent. Therefore, for each sensor we employ a sub-network with a LSTM layer to encode the temporal information, resulting in a network with $4$ parallel sub-LSTM networks. The output of these sub-networks are then concatenated and fed to a sequence of fully-connected layers with $512$, $256$, $128$, $64$, and $32$ neurons, respectively. Finally, a fully-connected layer with $2$ neurons is used to regress the two angles. We note that all the fully-connected layers use ReLU as the activation function, except the last layer uses the linear activation function.

Since all four sensors move simultaneously when the subject moves the shoulder, assuming that the input signals from these sensors are independent as in LNNet does not always hold in practice. In the second design (Early-fusion Nonlinearity Net - ENNet) (Fig. \ref{Fig_learning_overview}. b), we propose to use a unified LSTM network to handle all four input sensors simultaneously. The values of each time step of four sensors are considered as the features for the LSTM network. With this architecture, all the sensory data are learned simultaneously while the temporal information is effectively extracted by the LSTM. Similar to LNNet, we also use a series of full-connected layers with the ReLU as the activation function, and the final layer has two neurons to regress the angle values.

\textbf{Training}
We train both networks end-to-end using the mean squared error (MSE) $L_2$ loss function between the ground-truth angle values obtained from the optical tracking system, $y_i$ , and the predicted control from the network $\hat{y}$:

\begin{equation}
L(y,\hat{y}) = \frac{1}{m}\sum\limits_{i = 1}^m {({y_i} - {{\hat y}_i})^2}
\end{equation}

In practice, we train both networks end-to-end using Adam optimiser \cite{kingma2014adam} with the default parameters (i.e., learning rate, $\beta_1$, and $\beta_2$ are $0.001$, $0.9$, and $0.999$, respectively). The batch size and the number of LSTM steps are empirically set to $500$ and $10$, respectively. The networks are trained for $10,000$ epochs and the training time is approximately $3$ hours on an NVIDIA GTX 2080 GPU. The inference time of our networks is approximately $3ms$ on an NVIDIA GTX 2080 and $12ms$ on NVIDIA Jetson Nano, which allows us to do real-time control in many robotic applications.

\subsection{Simulated Teleoperation}

To validate the accuracy and efficacy of our  mapping, a framework to achieve real-time teleoperation was developed. The orientation representation was changed from the MoBL-ARMS MS model (in which the inverse kinematics was performed, see Section \ref{subsec:expMethod}) to that of the NAO model in V-REP (Virtual Robot Experimentation Platform, Coppelia Robotics, Switzerland). The orientation representation used in the MS model is a non-standard representation with angles measuring the plane of elevation $(elv\_angle (\theta))$, the shoulder elevation angle $(shoulder\_elv (\phi))$, and internal/external rotation $(shoulder\_rot)$. In V-REP, rotations are defined by Euler $X-Y-Z$ rotation sequence in the absolute (extrinsic) frame, or the Euler $Z-Y'-X''$ rotation sequence in the robot (intrinsic) frame. To obtain the equivalent orientation angles in the V-REP NAO representation from that of the MS model, the following steps were performed:
\begin{enumerate}
    \item The MS model representation was transformed to the $Y-X'-Y''$ intrinsic Euler rotations (International Society of Biomechanics convention) with rotation in $X'$ being negative in value, as described in \cite{xu2012coordinate}.
    \item The $Y-X'-Y''$ Euler angles are converted to equivalent quaternion representation values.
    \item The quaternion values were then converted to the $Z-Y'-X''$ intrinsic Euler rotations, to obtain equivalent representation for the NAO robot in V-REP.
\end{enumerate}
The conversion from $Y-X'-Y''$ to $Z-Y'-X''$ representation could have been performed directly, but was done using quaternions to leverage the advantages inherent in the representation. 

To implement teleoperation, the schematic presented in Fig.\ref{fig:teleopSchematic} was implemented. Data acquisition and processing from the four sensors was done in a separate thread running at a high frequency of $250\si{\hertz}$. In the main thread, sensor data was either called on-demand from the data acquisition thread or through pre-recorded sensor data running in a loop. This data was fed as input to the ENNet presented in Section \ref{subsec:learning}. The estimated joint angle value that is output from network is then communicated to the virtual robot model running in V-REP. The communication was implemented using the V-REP remote API in Python.

\section{EXPERIMENTAL RESULTS}
\label{sec:expRes}

\subsection{Learning Results}
We compare our learning results with the neural network method in \cite{varghese2019design}. Furthermore, we also compare our software-hardware based solution with the recent work in ~\cite{ogata2019estimating} which used the CNN to compensate the nonlinearity in the soft wearable suits, and the sensor-only solution using the IMU sensor in ~\cite{Aslani2018}. We notice that our proposed methods, \cite{ogata2019estimating}, and neural network \cite{varghese2019design} require training while the IMU solution in \cite{Aslani2018} can be directly used without the need for training.

Table \ref{tb_result} summarises the root mean square error (RMSE) of our proposed methods and the previous work in \cite{varghese2019design} since we use the same experimental setup. This table clearly shows that our ENNet achieves the lowest error in estimating both azimuth ($\theta$) and elevation ($\phi$). From table \ref{tb_result}, we notice that both our proposed methods LNNet and ENNet significantly outperform the simple neural network solution in ~\cite{varghese2019design}. This shows that learning the long-term dependencies in the sensor data is the key to success. Since the neural network method in \cite{varghese2019design} only employs a simple multi-layer neural network, it does not take into account the temporal dependency in the data as in our proposed methods. Overall, we also observe that our ENNet outperforms the LNNet, which shows that the early fusion strategy brings better results than the late fusion strategy. This result is understandable since in practice the movement of the shoulder joint affects all the sensors simultaneously, therefore the early fusion strategy would allow the ENNet to learn from all sensors more effectively than LNNet.

\begin{table}
\centering\ra{1.3}
\caption{The RMSE Error Over The Testing Set}
\label{tb_result}

\begin{tabular}{@{}rcccccc@{}}
\toprule 					&  
Training?     &
$\theta$ error          &
$\phi$ error      & 

\\
\midrule
ANN~\cite{varghese2019design} & Yes &  5.43  & 3.66    		  \\
LNNet (ours)  & Yes     & 3.07	 & 1.72    \\
ENNet (ours)  & Yes     & \textbf{2.51}  & \textbf{1.33}		   \\
\bottomrule
\hspace{2ex}

\end{tabular}
\end{table}

\begin{table}
\centering\ra{1.3}
\caption{Comparison of RMSE for Estimating 2 DoF Shoulder Joint Kinematics}
\label{tb_suit_comparision}

\begin{tabular}{@{}rcccccc@{}}
\toprule 					&  
Angle 1 error          &
Angle 2 error      & 

\\
\midrule
Xenoma \cite{nakajima2019wiring} + CNN \cite{ogata2019estimating} &  7.51  & 9.41    		  \\
IMU \cite{Aslani2018}      &3.00   & 2.00  		   \\
Sensing Suit \cite{varghese2019design} + ENNet (ours)   & \textbf{2.51}	 & \textbf{1.33}    \\
\bottomrule
\end{tabular}
\end{table}

Table \ref{tb_suit_comparision} provides a comparison between three wearable sensing technologies and their performance in estimating multi-DoF joint kinematics at the shoulder. IMU-based sensing technology is considered the benchmark for wearable kinematic sensing, and they do not need any learning-based estimation techniques to derive the joint kinematics. However, they suffer from drift which can increase uncontrollably if not corrected. This shortcoming is not encountered by the sensing suits presented in \cite{varghese2019design} and \cite{nakajima2019wiring}. The solution proposed in \cite{ogata2019estimating} used the Xenoma sensing suit \cite{nakajima2019wiring}, which is a commercially available high Technology Readiness Level (TRL) soft sensing suit based on fabric-based strain sensing. The sensing values were fed as input to a CNN to compensate nonlinearities and predict joint kinematics. The Xenoma suit's working principle can be considered to be equivalent to the working principle of the spline-generated virtual sensing framework in \cite{varghese2019design} as they both compute strains from fabric stretch. However, it can be seen that even though the Xenoma suit is of a higher TRL than the sensing suit presented in \cite{varghese2019design}, the CNN does not learn the nonlinearity compensation well and the RMSE of the Xenoma+CNN for multi-DoF kinematics estimation is much poorer as compared to the IMU and the solution presented in this work. It is however to be noted that the orientation representation used in \cite{ogata2019estimating} is different from the representation used in this work. Two of the relevant shoulder DoF angles with the lowest error were hence chosen for comparison in Table \ref{tb_suit_comparision}. It can also be seen that sensing suit \cite{varghese2019design} with the spline-generated sensing framework + ENNet nonlinearity compensator is able to learn the rate-dependent total nonlinearity and outperform even the drift-prone state-of-the-art IMU-based solution. With further development, this solution could be made even more robust and potentially replace an IMU-based kinematic sensing system for exosuit control and other applications.


\begin{figure}[t]
\centering
\includegraphics[width = \linewidth]{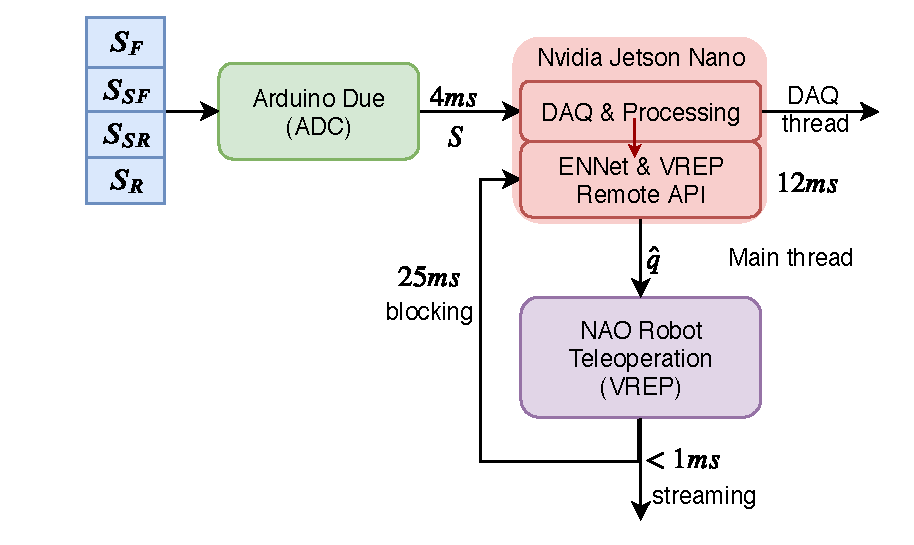}
\caption{Overview of flow of information for teleoperation and computational performance. The flow of information happens over two threads running on the Jetson Nano. One thread runs the sensor data acquisition from the Arduino Due, and the second (main) thread performs the joint kinematics prediction and transmits it to the robot model in V-REP}
\label{fig:teleopSchematic}
\end{figure}

\begin{figure}[t]
\centering
\includegraphics[width=\linewidth]{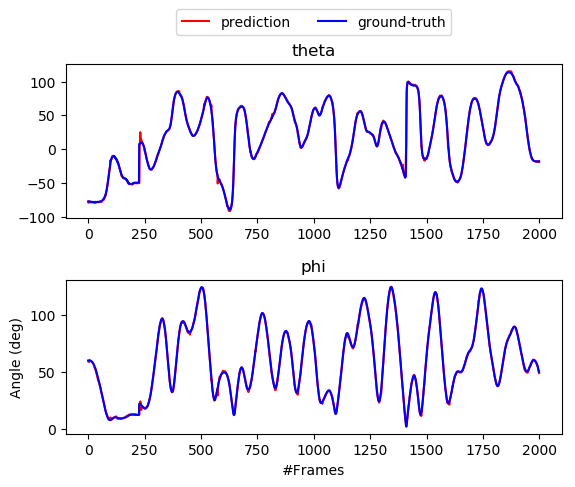}

\caption{Comparison of joint-angle prediction by ENNet against ground-truth for 2000 frames of random movements while performing teleoperation in simulation.}
\label{fig_prediction_result}
\end{figure}

\subsection{Teleoperation Results}
\label{teleop}
The aim of the kinematic sensing framework is to be part of a sensing network with an aim to provide robust exosuit control and context-based intention-detection to the user. To this end, we validated the effectiveness and performance of the developed learning-based nonlinearity compensation framework in being able to accurately compensate nonlinearities and predict joint kinematics in real-time. The trained network was deployed on an edge-computing device, the Jetson Nano. The Jetson Nano has a 128-core Maxwell GPU and was used to run the trained network for predicting joint kinematics from the 4 sensor values. An overview of the information flow in the implementation of the teleoperation is shown in Fig.\ref{fig:teleopSchematic}. Computation times for entities of the teleoperation setup are presented in Fig.\ref{fig:teleopSchematic} as well. The data acquisition thread runs separately on the Jetson Nano at $250\si{\hertz}$. The prediction/inference computation on the ENNet takes approximately $12\si{\milli}\si{\second}$. The joint kinematics were transmitted to the NAO robot in V-REP using the V-REP Remote API for Python. Using the remote API with V-REP running on a different computer, transmitting one set of joint angles took approximately $25\si{\milli}\si{\second}$ in blocking mode (ensuring the every movement is executed), or $<1\si{\milli}\si{\second}$ in streaming mode (non-blocking mode). The learned network was able to achieve excellent prediction results in tracking joint kinematics in 2 DoFs. A snapshot of teleoperation performed over joint angle data predicted from 2000 frames of spline-generated sensor values, and the corresponding ground-truth data is presented in Fig.\ref{fig_prediction_result}.

\section{CONCLUSION}
\label{sec:conc}
Soft wearable kinematic sensing frameworks are vital for exosuits to be able to provide intuitive and natural assist-as-needed control to a user. Apart from exosuit control, wearable sensing frameworks could see applications ranging from VR/AR to continuous health monitoring. However, the soft and compliant nature of exosuits introduce highly coupled rate-independent (hysteresis) and rate-dependent (creep) nonlinearities within both the actuation and sensing framework that are difficult to isolate and eliminate. For the sensing framework to act as a robust benchmark for the actuation system of a robot, these inherent nonlinearities need to be estimated and corrected. In this work, we introduced a new deep learning-based sensor fusion and nonlinearity compensation framework. The LSTM-based framework encodes both temporal and spatial information and is able to successfully compensate the rate-dependent total nonlinearity present in the sensing suit, and accurately predict joint kinematics. The spline-generated (virtual) sensing framework and the ENNet nonlinearity compensation presented here was able to outperform the drift-prone state-of-the-art IMU-based solution. The virtual spline-based sensing was used so that the proposed nonlinearity compensation is not affected by the physical limits that were imposed on the real sensing system in \cite{varghese2019design}. We further validated the effectiveness of the developed framework by performing real-time simulated teleoperation using pre-recorded sensor values with a NAO humanoid robot. The sensing suit and ENNet nonlinearity compensation achieved a tracking frequency of over $25\si{\hertz}$, making the presented solution suitable for real-time exosuit control and teleoperation. In the future, we intend to generalise this framework to be able to automatically adapt to multiple users and multiple donning/doffing trials with minimal training and human intervention. Additionally, we intend to further improve this work by compensating and correcting for both rate-dependent (creep) and rate-independent (hysteresis) independently, and potentially achieve more accurate joint kinematics estimation.

\section*{ACKNOWLEDGMENT}
The authors would like to acknowledge the EPSRC, UK for funding this work. The authors would also like to thank Dr. Yao Guo, Daniel Freer, Ya-Yen Tsai and Daniel Bautista for their help with motion-capture experiments.

\bibliographystyle{IEEEtran}
\bibliography{IEEEabrv,RoboSoft2020_Hysteresis_References}

\end{document}